# Coupling Symbolic Reasoning with Language Modeling for Efficient Longitudinal Understanding of Unstructured Electronic Medical Records

Shivani Shekhar*
Mendel Health Inc.
California, USA
shivani.s@mendel.ai

Simran Tiwari*
Mendel Health Inc.
California, USA
simran.t@mendel.ai

T. C. Rensink
Mendel Health Inc.
California, USA
thomas.r@mendel.ai

Ramy Eskander
Mendel Health Inc.
New York, USA
ramy.e@mendel.ai

Wael Salloum
Mendel Health Inc.
California, USA
wael@mendel.ai

## ABSTRACT

The application of Artificial Intelligence (AI) in healthcare has been revolutionary, especially with the recent advancements in transformer-based Large Language Models (LLMs). However, the task of understanding unstructured electronic medical records remains a challenge given the nature of the records (e.g., disorganization, inconsistency, and redundancy) and the inability of LLMs to derive reasoning paradigms that allow for comprehensive understanding of medical variables. In this work, we examine the power of coupling symbolic reasoning with language modeling toward improved understanding of unstructured clinical texts. We show that such a combination improves the extraction of several medical variables from unstructured records. In addition, we show that the state-of-the-art commercially-free LLMs enjoy retrieval capabilities comparable to those provided by their commercial counterparts. Finally, we elaborate on the need for LLM steering through the application of symbolic reasoning as the exclusive use of LLMs results in the lowest performance.

## CCS CONCEPTS

• **Computing methodologies** → **Artificial intelligence**; • **Applied computing** → *Health informatics*.

## KEYWORDS

AI in Healthcare, Symbolic Reasoning, Generative Language Models, Natural Language Processing, Unstructured Electronic Medical Records

## 1 INTRODUCTION

Clinical language is convoluted as patient records often contain discrepancies that require human expertise to decode. For instance, *Seizures* can either be a symptom or a side effect, while a pathology report indicating cancer of some stage might conflict with a doctor's note indicating a different one. In addition, patient records are often unstructured and involve materials irrelevant to the patient journey; examples include routine forms and generic medical references, such as a listing of possible side effects. While Artificial Intelligence (AI) has been shaping a pivotal shift in the healthcare industry [30, 32], the aforementioned challenges limit the capabilities of AI in clinical abstraction.

Transformer-based Large Language Models (LLMs), autoregressive generative models in particular, have been revolutionizing the role of AI in several industries, one of which is healthcare [4, 21, 31]. LLMs have proved exceptional capabilities at both retrieval and generation, where they might replace whole machine learning pipelines of complex architectures. Yet, generative LLMs surprisingly fail at solving trivial tasks. This is mainly due to the fact that those models tend to memorize sequences of operations in a shallow learning manner rather than composing proper reasoning paradigms that reflect holistic task understanding [6, 10, 36].

In this work, we examine the power of coupling symbolic reasoning (symbolic AI) with language modeling to better utilize the retrieval and generation capabilities of LLMs towards an improved understanding of unstructured electronic medical records. We achieve this by leveraging state-of-the-art LLMs within a clinical abstraction pipeline whose core understanding component utilizes both Natural Language Processing (NLP) and symbolic reasoning, denoted by NLP_REASONING. We compare NLP_REASONING to the following three setups:

(1) RET_LLM_NLP_REASONING: An LLM-based setup that employs NLP_REASONING on top of chunks retrieved by retrieval LLMs.
(2) GEN_LLM_NLP_REASONING: An LLM-based setup that employs NLP_REASONING on top of answers retrieved by generative LLMs.
(3) RET_GEN_LLM_NLP_REASONING: An LLM-based setup that employs NLP_REASONING on top of chunks and answers retrieved by retrieval and generative LLMs.

In addition, we conduct an additional experiment in which a generative LLM is utilized as an end-to-end solution that replaces NLP_REASONING.

Our contribution in this paper is threefold:

(1) We assess the power of coupling symbolic reasoning with language modeling for the longitudinal understanding of unstructured electronic medical records.
(2) We compare several state-of-the-art LLMs in terms of their retrieval and generation capabilities.

---
*These two authors contributed equally to the work.

(3) We manually assess the performance of generative LLMs as a standalone system that replaces an established architecture that leverages symbolic reasoning.

Evaluating our abstraction pipeline on the extraction of 13 cancer-related medical variables from unstructured medical records shows that the best solution is one that combines symbolic reasoning with language modeling, where the understanding of several medical variables benefits from the retrieval and generation capabilities of LLMs. In addition, applying symbolic reasoning on chunks and answers produced by LLMs instead of full records decreases compute cost and time. We also show that the state-of-the-art commercially-free retrieval LLMs perform as well as their commercial counterparts. At last, we highlight the necessity of symbolic reasoning as an LLM steering mechanism as solely relying on generative LLMs for medical abstraction significantly lowers the performance .

## 2 RELATED WORK

Over the past decade, Artificial Intelligence (AI) has proved successful in healthcare through an enormous number of applications [28] that save enormous clinical work and improve patient journey. Example applications include prediction of disease outbreaks, such as Malaria [35] and COVID-19 [14], prediction of diseases through genomic analyses [9, 22], improvement of mental health [11, 16], drug discovery [5, 7, 25, 29], AI-based chatbots in healthcare [2, 8, 13, 27] and disease diagnosis [20, 33], especially for cancer [12, 17–19].

Applying AI for the processing of electronic medical records has been receiving increasing interest, especially in the case of unstructured records [15, 23, 24]. Such records, while containing valuable information, remain difficult to process, which necessitates the use of AI for the segmentation, classification and parsing of the documents prior to extracting the valuable information and ingesting them into downstream applications. In this paper, we discuss how to leverage NLP, symbolic reasoning and LLMs for the longitudinal understanding of unstructured clinical texts.

AI in healthcare is now attracting a bigger community given the recent revolutionary advancements in transformer-based Large Language Models (LLMs) [4, 21, 31], especially autoregressive generative models, such as the GPT-family models [3] and PaLM [1]. Such widespread interest is attributed to the current commoditization of such models, along with the ability to pursue customization through fine-tuning with custom data. However, LLMs come with limitations as they are unable to process complex logic and perform robust generalization, which hinders the proper understanding and processing of complex inputs [6, 10, 36]. Accordingly, we assess the use of LLMs in conjunction with symbolic reasoning towards better understanding of challenging clinical texts.

## 3 METHODS

The input to our clinical abstraction pipeline is a set of unstructured electronic medical records that belong to some patient and come in image and PDF formats, while our output is structured medical variables describing the patient journey.

Next, we describe our clinical abstraction pipeline (See Appendix A.1). This includes the preprocessing of the unstructured medical records and our core abstraction system that leverages both NLP and symbolic reasoning, denoted by NLP_REASONING. We then describe three setups that utilize LLMs for both retrieval and generation within our pipeline.

### 3.1 Preprocessing

*3.1.1 Optical Character Recognition (OCR).* We utilize an in-house OCR system to convert the input images of medical records into texts ready to be ingested for processing.

*3.1.2 De-identification.* We employ a transformer-based sequence tagging architecture for the redaction of Protected Health Information (PHI) in the texts extracted by the OCR module to comply with HIPPA security rules [1].

### 3.2 Medical Abstraction: NLP_REASONING

*3.2.1 Document Segmentation and Segment Classification.* The output of the de-identification module is in the form of text files that may belong to multiple clinical documents. Accordingly, we perform document segmentation by leveraging an ensemble of embedding-based and rule-based approaches. We then classify the documents into their proper categories, e.g., *Pathology*, *Administrative*, *Lab Results*, *SOAP Note*, etc.

*3.2.2 Entity and Relation Extraction.* We employ a transformer-based sequence tagging architecture for the tagging of predefined medical entities, e.g., *Cancer* and *Surgery*. A transformer-based context classification model then determines the relation between every possible pair of extracted entities, if any. The outputs are then stored in tag graphs in which the nodes and edges represent the entities and their relations, respectively.

*3.2.3 Symbolic Reasoning.* We utilize an in-house proprietary medical ontology that is a comprehensive, hierarchical representation of medical terminology and knowledge. It contains an expansive set of medical concepts and the relations between them.

Our symbolic reasoning operates on document-level and patient-level bases. On the document level, we join the underlying tag graphs with the ontology to extract an object graph that represents the underlying document. This is conducted with the help of rule-based approaches along with voting mechanisms to resolve disagreements. The object graph captures all the medical concepts present in the document and the relations between them.

Next, the patient-level processing consolidates the document-level object graphs to form a heterogeneous patient-level graph. The graph is formed by combining those objects that are compatible with one another through the application of a large set of logical rules that traverse the ontology, along with confidence assignments that represent the likelihood of the objects. For instance, a *Neoplasm* node consolidates several *Neoplasm* nodes that are consistent with one another; a Neoplasm node for *Cancer*, for example, is consistent with another Neoplasm node for *Lung Cancer* since *Lung Cancer* is a subtype of *Cancer*, while *Breast cancer* and *Lung Cancer*, on the other hand, are incompatible since they do not refer to the same type of *Cancer*. Such compatibility rules are traced transitively across the ontology.

---

[1] https://www.hipaajournal.com/considered-phi-hipaa/

Coupling Symbolic Reasoning with Language Modeling for Efficient Longitudinal Understanding of Unstructured Electronic Medical Records## 3.3 Integration of LLMs

We integrate LLMs into our abstraction pipeline by querying the LLMs with the output of the preprocessing steps and passing the retrieved and/or generated outputs to our core abstraction system NLP_REASONING. We describe our LLM-based setups below.

*3.3.1 LLMs for Retrieval.* In this setup, we apply retrieval (embedding) LLMs on the preprocessed documents along with a predefined set of 31 questions (See Appendix A.2) that targets a predefined set of 13 cancer-related medical variables that we need to extract (See Appendix A.3). For each question, the LLMs retrieve four semantically related chunks from the preprocessed documents. We then combine all the retrieved chunks and pass them as the sole input to NLP_REASONING (as opposed to passing all the preprocessed documents). We refer to this setup as RET_NLP_REASONING.

*3.3.2 LLMs for Generation.* In this setup, we pass the retrieved chunks from RET_NLP_REASONING along with the corresponding questions as input to a generative LLM to obtain an answer for each question. We then combine all the answers and pass them as the sole input to NLP_REASONING. We refer to this setup as GEN_NLP_REASONING.

*3.3.3 LLMs for Retrieval and Generation.* This setup combines RET_NLP_REASONING and GEN_NLP_REASONING by combining both the retrieved chunks and the generated answers and passing them as the sole input to NLP_REASONING. We refer to this setup as RET_GEN_NLP_REASONING.

*3.3.4 Standalone LLMs.* Finally, we leverage LLMs as a standalone system in the placement of NLP_REASONING, where the answers returned by the generative model are the final outcome of the abstraction pipeline.

## 4 EVALUATION AND ANALYSIS
## 4.1 Experimental Settings

*4.1.1 Data.* We conduct our experiments for the extraction of 13 cancer-related medical variables, namely *Neoplasm*, *Morphology*, *T-Stage*, *N-Stage*, *M-Stage*, *Stage Group*, *Medications*, *Outcome*, *Response*, *Tested Biomarkers*, *Surgeries*, *Diagnostic Procedures* and *Cancer Diagnosis Date* (See Appendix A.3 for the descriptions). We process 100 patients of three cancer types, namely *Colorectal Cancer* (42 patients), *Breast Cancer* (28 patients), and Lung Cancer (30 patients); the patients are uniformly sampled with respect to the number of corresponding records, which varies between 5 and 75, with an average of 34 records per patient.

*4.1.2 Language Models.* We conduct our experiments using a number of state-of-the-art LLMs by leveraging their retrieval and generation capabilities. The models and their settings are listed in Table 1. The retrieval chunk sizes are selected such that the combined size of the input chunks, prompts, and answers does not exceed the maximum token limit of the underlying LLM and the potential of our compute resources, while the temperatures are optimized for abstraction quality. See Appendix A.2 for the complete set of questions.

| Language Model | Retrieval | Generation |
|---|---|---|
| GPT3.5 [2] | chunk_size=3k | temperature=0.67 |
| ada2 [3] | chunk_size=1,612 | n/a |
| PaLM2 [1] | chunk_size=3k | temperature=0.67 |
| Instructor-XL [4] [34] | chunk_size=3k | n/a |
| SGPT [5] [26] | chunk_size=1,612 | n/a |
| Dolly2-12b [6] | n/a | temperature=1.0 [7] |
| Falcon-7b-Instruct [8] | chunk_size=1,612 | temperature=1.0 |

Table 1: Utilized LLMs and their Settings

*4.1.3 Performance Evaluation.* We evaluate the performance of our setups, namely NLP_REASONING, RET_NLP_REASONING, GEN_NLP_REASONING and RET_GEN_NLP_REASONING, using in-house manually annotated gold standard for the aforementioned medical variables in terms of F1-score, precision and recall.

## 4.2 System Performance
The abstraction performance of our baseline (NLP_REASONING) and the three LLM-based setups that couple symbolic reasoning with language modeling is reported in Table 2.

## 4.3 Analysis
*4.3.1 Abstraction of Medical Variables.* There is vast variation among the medical variables in how the different setups perform and which is best. The integration of LLMs achieves the biggest improvements at the detection of *Surgeries* (GEN_NLP_REASONING with GPT3.5), *Response* (GEN_NLP_REASONING with GPT3.5) and *Cancer Diagnosis Date* (GEN_NLP_REASONING with PaLM2), with relative F1-score improvements of 35.31%, 28.97% and 16.33%, respectively. Overall, LLMs help in the cases of subjective variables, such as *Response*, due to their ability to understand the semantics, and in the cases of variables of multiple values, such as *Medications* and *Surgeries*, due to their ability to associate related items together. On the other hand, LLMs do not help in the abstraction of *Neoplasm*, due to the high baseline provided by NLP_REASONING, while they suffer at abstracting variables that are scattered across the record, such as *Tested Biomarkers*.

*4.3.2 Retrieval vs. Generation.* The overall performance indicates that retrieval models accurately select text chunks with relevant information to the asked questions, and thus improve the abstraction of several variables, such as *T-Stage*, *Stage Group*, *Medications* and *Outcome*. Generated answers, on the other hand, are less beneficial to the core reasoning system except in the abstraction of *Response*, *Surgeries* and *Cancer Diagnosis Date*, at which NLP_REASONING results in low precisions (See Table 4). This suggests that the reasoning system gets overwhelmed by large amounts of information for these variables, increasing the number of false positives, while

---
[2] https://platform.openai.com/docs/models/gpt-3-5
[3] https://platform.openai.com/docs/models/embeddings
[4] https://huggingface.co/hkunlp/instructor-xl
[5] https://github.com/Muennighoff/sgpt
[6] https://huggingface.co/databricks/dolly-v2-12b
[7] In the case of Dolly2-12b, we use Instructor-XL to generate the embeddings.
[8] https://huggingface.co/tiiuae/falcon-7b-instruct

Shivani Shekhar*, Simran Tiwari*, T. C. Rensink, Ramy Eskander, and Wael Salloum

| LLMs | Neoplasm | Morphology | T-Stage | N-Stage | M-Stage | Stage Group | Medications | Outcome | Response | Tested Biomarkers | Surgeries | Diagnostic Procedures | Cancer Diagnosis Date |
|---|---|---|---|---|---|---|---|---|---|---|---|---|---|
| | | | | | NLP_REASONING | | | | | | | | |
| None | **92.9** | 80.5 | 80.7 | 73.2 | 78.0 | 68.8 | 48.8 | 45.8 | 45.6 | **70.5** | 36.6 | **68.0** | 60.3 |
| | | | | | RET_NLP_REASONING | | | | | | | | |
| GPT3.5 | 91.88 | 82.35 | 81.09 | 76.86 | 76.61 | **75.04** | 48.18 | 43.52 | 48.73 | 69.82 | 36.96 | 66.26 | 61.04 |
| ada2 | 91.78 | 80.21 | 76.87 | 76.12 | 75.05 | 67.71 | 48.52 | 47.82 | 45.20 | 65.46 | 37.65 | 66.75 | 55.54 |
| PaLM2 | 89.54 | 81.12 | 83.59 | 75.83 | 73.89 | 68.11 | **52.79** | 40.48 | 47.90 | 68.79 | 39.43 | 64.39 | 63.73 |
| Instructor-XL | 91.88 | 82.14 | **83.81** | 75.83 | 72.73 | 67.07 | 51.26 | **48.37** | 45.00 | 67.37 | 36.61 | 66.83 | 56.68 |
| SGPT | 91.69 | 76.89 | 77.45 | 70.84 | 69.16 | 66.92 | 48.37 | 42.25 | 46.22 | 61.39 | 37.36 | 64.60 | 60.21 |
| Falcon-7b-Instruct | 90.00 | 77.53 | 81.02 | 65.31 | 59.79 | 61.70 | 44.99 | 28.57 | 47.16 | 62.30 | 37.63 | 57.43 | 59.79 |
| | | | | | GEN_NLP_REASONING | | | | | | | | |
| GPT3.5 | 88.63 | 80.00 | 60.35 | 58.91 | 57.33 | 56.04 | 45.79 | 19.23 | **58.76** | 50.48 | **49.51** | 46.84 | 53.58 |
| PaLM2 | 77.46 | 74.32 | 61.64 | 60.98 | 50.18 | 45.71 | 46.27 | 27.24 | 52.17 | 49.44 | 48.77 | 33.18 | **70.16** |
| Dolly2-12b | 61.32 | 60.32 | 26.94 | 29.68 | 15.16 | 30.15 | 22.54 | 22.04 | 41.50 | 27.31 | 23.25 | 34.76 | 21.45 |
| Falcon-7b-Instruct | 53.88 | 11.38 | 2.40 | 0.00 | 2.81 | 10.79 | 14.23 | 14.29 | 51.30 | 4.43 | 23.88 | 35.28 | 13.33 |
| | | | | | RET_GEN_NLP_REASONING | | | | | | | | |
| GPT3.5 | 90.86 | **84.29** | 80.54 | **78.14** | 76.61 | 73.01 | 47.70 | 43.04 | 46.47 | 68.52 | 37.37 | 67.55 | 61.97 |
| PaLM2 | 83.96 | 83.06 | 83.72 | 76.86 | 75.09 | 69.15 | 50.00 | 40.68 | 48.51 | 66.37 | 37.96 | 64.33 | 57.93 |
| Dolly2-12b | 90.15 | 83.27 | 81.76 | 74.57 | **78.6** | 58.96 | 40.97 | 38.76 | 38.08 | 54.06 | 27.04 | 62.24 | 36.37 |
| Falcon-7b-Instruct | 89.18 | 77.53 | 81.02 | 65.31 | 59.79 | 60.72 | 42.11 | 28.21 | 46.41 | 61.95 | 31.52 | 56.64 | 50.31 |

Table 2: System Performance (F1-Score)%. The best result per medical variable is in bold.

generating synthetic responses is valuable in such cases. The generated answers, however, appear to confuse the reasoning system when combined with the relevant chunks except in the cases of *Morphology*, *N-stage*, and *M-Stage*.

*4.3.3 Comparison of LLMs.* The performance of the retrieval models does not noticeably vary, only a maximum difference of 3.46% in macro-averaged F1-score across the medical variables, when excluding the least performing model Falcon-7b-Instruct. This suggests that some state-of-the-art commercially-free models (ada2, Instructor-XL, and SGPT) provide performance comparable to the commercial ones (GPT3.5 and PaLM2) at hand at the time of writing this paper. Although generated answers are less efficient for this task, GPT3.5 outperforms PaLM2, while the less restricted Dolly2-12b and Falcon-7b-Instruct fall behind in performance, with macro-averaged F1-scores of 55.80%, 53.66%, 32.03%, and 18.31%, respectively, across all the medical variables.

*4.3.4 Anomalies of LLMs.* One noticeable behavior is that the responses generated by the LLMs exhibit inconsistencies. For instance, a patient summary provided by PaLM2 (in response to Question 1 in Appendix A.2) states that the patient has *Breast Cancer*, whereas it states the patient has *Lung Cancer* in response to the subsequent question about the cancer type. Similar issues can be observed with the use of GPT3.5, Dolly2-12b, and Falcon-7b-Instruct as well.

Moreover, LLMs might confuse patient-related information with generic information, e.g., confusing the list of medications the patient had with a recommended list of medications. In addition, generative LLMs might suffer from hallucinations when the targeted answer is not obvious in the underlying text. These two factors highly impact the performance of both GEN_NLP_REASONING and RET_GEN_NLP_REASONING.

### 4.4 LLMs as Standalone Systems

Next, we experiment with GPT3.5 as an end-to-end abstraction system without employing NLP_Reasoning and manually compare the generated answers to our gold standard. The exclusive use of GPT3.5 results in a relative decrease of 64.72% in the average F1-score achieved by NLP_Reasoning across the 13 medical variables, where GPT3.5 performs comparably to NLP_Reasoning only in the cases of objective and single-valued variables, such as the neoplasm and staging information. This is basically due to the anomalies discussed in Section 4.4. The results highlight the need to introduce symbolic reasoning to better exploit the generative capabilities of language modeling. In addition, the generated answers do not form structured outputs ready for ingestion into downstream applications, which requires additional layers of abstraction.

## 5 CONCLUSION AND FUTURE WORK

In this paper, we assessed the performance of coupling symbolic reasoning with language modeling toward a better understanding of unstructured electronic medical records. We evaluated our setups on 13 cancer-related medical variables and showed that the retrieval and generation capabilities of LLMs improve the abstraction of the majority of the variables. This combination also allows for lower compute cost and time since our core abstraction system is applied on chunks and answers produced by LLMs instead of full records. We also showed that commercially-free retrieval LLMs perform comparably to their commercial counterparts. Finally, we showed that the exclusive use of generative LLMs considerably drops the overall performance, which elaborates the need to introduce symbolic reasoning as an LLM steering mechanism.

In the future, we plan to introduce our in-house language models that are tailored for the understating of medical records towards enhanced abstraction of medical variables.

# A APPENDIX

## A.1 Abstraction Pipeline

Figure 1 depicts our pipeline before the integration of LLMs.

## A.2 Questions

We list below the set of questions used by the LLMs to retrieve the chunks and generate the answers for our 13 medical variables. The questions are a mix of simple and compound ones that map to one or more endpoints and are tuned to ensure reasonable outcomes.

(1) Summarize the patient.
(2) What cancer does the patient have?
(3) What is the morphology of the cancer?
(4) What are the T, N and M stages of the cancer?
(5) What is the stage group of the cancer?
(6) What is the date of finding the T, N and M stages of the cancer?
(7) What is the date of diagnosis of the cancer?
(8) What is the date of diagnosing the metastasis of the cancer?
(9) Did the patient undergo radiation therapy?
(10) List all the cancer-related medications that the patient took.
(11) List all the medications the patient used for cancer and their start dates.
(12) On what date did the patient start taking each medication?

Shivani Shekhar*, Simran Tiwari*, T. C. Rensink, Ramy Eskander, and Wael Salloum

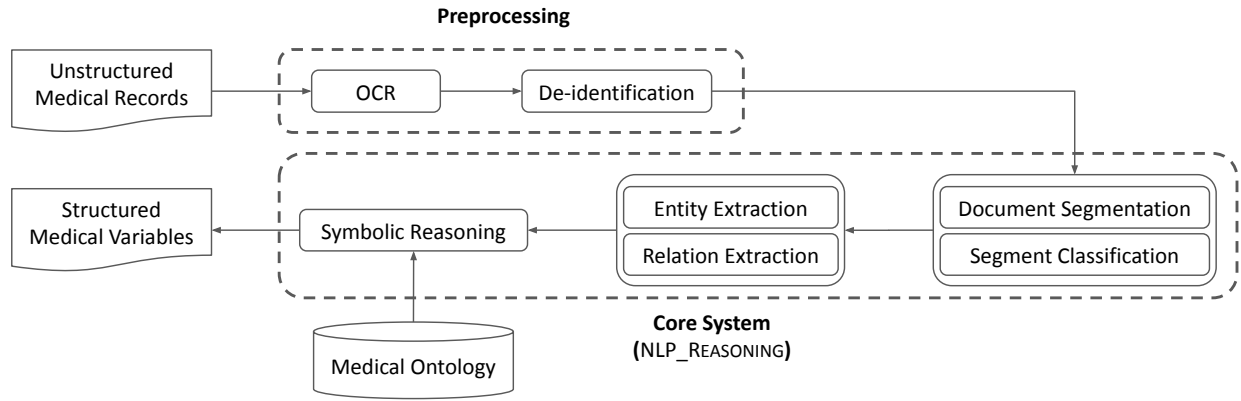

Figure 1: Abstraction Pipeline

(13) List all the cancer-related surgeries that the patient underwent.
(14) What are the dates of all the cancer-related surgeries that the patient underwent?
(15) List all the cancer-related surgeries that the patient underwent and their dates.
(16) What is the current status of the cancer?
(17) What is the current status of the cancer and the date of diagnosis of the outcome?
(18) What was the response of the patient to the cancer treatment?
(19) How did the patient respond to the cancer-related medications and treatments, and what are the dates of the response?
(20) What is the date of diagnosing the patient's response to the cancer-related medications and treatments?
(21) List all the biomarkers that the patient was tested for.
(22) List all the biomarkers that the patient was tested for and their interpretations.
(23) List all the interpretations of each biomarker that the patient was tested for.
(24) List all the biomarkers that the patient was tested for and their categorical values.
(25) List the categorical values of all the biomarkers that the patient was tested for.
(26) List all the biomarkers that the patient was tested for and their methods of testing.
(27) List all the methods of testing of each biomarker that the patient was tested for.
(28) List all the cancer-related diagnostic procedures that the patient underwent.
(29) What are the dates of all the cancer-related procedures that the patient underwent?
(30) What are the dates of all the cancer-related diagnostic procedures that the patient underwent?
(31) List all the cancer-related diagnostic procedures that the patient underwent and their dates.

### A.3 Medical Variables

Table 3 lists the 13 cancer-related medical variables that we abstract from the medical records.

| Medical Variable | Description |
| --- | --- |
| Neoplasm | The patient's current cancer diagnosis (primary topography site) |
| Morphology | The description of the histological type present in a pathology specimen or recorded by a physician for a patient's cancer |
| T-Stage | The size and location of the tumor, including how much the tumor has grown into nearby tissues |
| N-Stage | The number of lymph nodes that have cancer |
| M-Stage | The distant metastatic status of the primary cancer |
| Neoplasm Stage Group | An assessment based on the combined evaluation of the patient's T stage, M stage, and N stage and prognostic parameters for specific cancers |
| Medications | The name of the chemotherapy, targeted therapy, endocrine therapy, or immunotherapy received by the patient for the current cancer diagnosis |
| Outcome | A physician-reported outcome attributable to local treatment, e.g., surgery and radiotherapy, or systemic treatment, e.g., chemotherapy (Values: Remission, Recurrence, etc.) |
| Response | A physician-reported assessment of the effectiveness of a therapeutic procedure (Values: Disease Progression, Stable Response, etc.) |
| Tested Biomarkers | Tests to check for specific biomarkers |
| Surgeries | A list of surgeries that the patient underwent |
| Diagnostic Procedures | Tests related to the patient's cancer diagnosis |
| Cancer Diagnosis Date | The earliest pathologic evidence of the patient's current cancer diagnosis |

Table 3: Medical Variables and their Definitions

### A.4 Precision and Recall of System Performance

The abstraction performance of our baseline (NLP_REASONING) and the three LLM-based setups that couple symbolic reasoning with language modeling is reported in Table 4 and Table 5 for Precision and Recall, respectively.

| LLMs | Neoplasm | Morphology | T-Stage | N-Stage | M-Stage | Stage Group | Medications | Outcome | Response | Tested Biomarkers | Surgeries | Diagnostic Procedures | Cancer Diagnosis Date |
|---|---|---|---|---|---|---|---|---|---|---|---|---|---|
| | | | | | NLP_Reasoning | | | | | | | | |
| **None** | **91.50** | 78.90 | 77.53 | 71.69 | **75.26** | 66.50 | 36.73 | 47.38 | 37.27 | 76.09 | 24.69 | 81.40 | 59.39 |
| | | | | | RET_NLP_Reasoning | | | | | | | | |
| **GPT3.5** | 90.50 | 80.70 | 78.42 | 74.72 | 72.71 | 71.97 | 36.06 | 43.04 | 39.87 | 76.64 | 24.95 | 79.94 | 60.10 |
| **ada2** | 90.40 | 78.99 | 74.34 | 77.27 | 73.04 | 69.62 | 37.10 | 49.52 | 37.57 | 75.57 | 26.12 | 82.26 | 54.69 |
| **PaLM2** | 88.20 | 79.50 | 81.89 | 74.23 | 72.55 | 68.73 | 40.73 | 43.59 | 39.60 | 76.73 | 27.67 | 81.47 | 62.76 |
| **Instructor-XL** | 90.50 | 80.50 | 81.05 | 74.23 | 70.18 | 64.83 | 39.10 | 50.73 | 37.20 | 76.23 | 25.12 | 81.19 | 55.82 |
| **SGPT** | 91.22 | 76.49 | 78.00 | 73.65 | 68.52 | **72.50** | 37.63 | 48.29 | 39.29 | 76.91 | 27.15 | 84.75 | 59.90 |
| **Falcon-7b** | 89.09 | 76.73 | 84.09 | 69.67 | 66.74 | 65.40 | 37.31 | 40 | 42.83 | **78.29** | 27.98 | 81.81 | 59.18 |
| | | | | | GEN_NLP_Reasoning | | | | | | | | |
| **GPT3.5** | 87.30 | 80.42 | 81.19 | **79.02** | 69.73 | 60.61 | **43.19** | **71.43** | **47.38** | 74.02 | 42.18 | **93.11** | 52.76 |
| **PaLM2** | 76.30 | 75.11 | 73.73 | 68.89 | 48.42 | 41.14 | 39.08 | 60.77 | 39.68 | 68.64 | **42.80** | 92.29 | **69.08** |
| **Dolly2-12b** | 60.40 | 61.65 | 47.14 | 52.22 | 17.69 | 26.53 | 14.86 | 20.38 | 31.12 | 29.28 | 16.09 | 59.64 | 21.12 |
| **Falcon-7b** | 53.33 | 21.76 | 22.50 | 0.00 | 8.18 | 14.55 | 14.02 | 20.00 | 39.78 | 55.17 | 22.60 | 52.12 | 13.20 |
| | | | | | RET_GEN_NLP_Reasoning | | | | | | | | |
| **GPT3.5** | 89.50 | **82.60** | 77.40 | 75.97 | 72.71 | 67.01 | 35.52 | 42.13 | 36.32 | 71.25 | 25.01 | 80.00 | 61.02 |
| **PaLM** | 82.70 | 81.40 | 82.03 | 74.72 | 70.16 | 65.81 | 37.65 | 41.63 | 38.24 | 68.03 | 26.23 | 80.31 | 57.04 |
| **Dolly2-12b** | 88.80 | 81.60 | 78.57 | 72.50 | 73.44 | 50.38 | 28.64 | 32.21 | 28.66 | 47.24 | 16.79 | 70.45 | 35.82 |
| **Falcon-7b** | 88.28 | 76.73 | **84.09** | 69.67 | 66.74 | 61.27 | 32.92 | 33.33 | 37.07 | 76.55 | 21.72 | 65.48 | 49.79 |

Table 4: System Performance (Precision)%. The best result per medical variable is in bold.

| LLMs | Neoplasm | Morphology | T-Stage | N-Stage | M-Stage | Stage Group | Medications | Outcome | Response | Tested Biomarkers | Surgeries | Diagnostic Procedures | Cancer Diagnosis Date |
|---|---|---|---|---|---|---|---|---|---|---|---|---|---|
| | | | | | NLP_Reasoning | | | | | | | | |
| **None** | **94.33** | 82.19 | 84.08 | 74.85 | 80.94 | 71.25 | 72.56 | 44.22 | 58.57 | 65.66 | 70.66 | 58.46 | 61.26 |
| | | | | | RET_NLP_Reasoning | | | | | | | | |
| **GPT3.5** | 93.30 | 84.06 | 83.94 | 79.12 | 80.94 | 78.39 | 72.56 | 44.00 | 62.65 | 64.11 | 71.25 | 56.58 | 62.00 |
| **ada2** | 93.20 | 81.46 | 79.58 | 75.00 | 77.17 | 65.89 | 70.12 | 46.22 | 56.73 | 57.74 | 67.37 | 56.16 | 56.42 |
| **PaLM2** | 90.93 | 82.81 | 85.35 | 77.50 | 75.28 | 67.50 | **75.00** | 37.78 | 60.61 | 62.34 | 68.62 | 53.24 | 64.74 |
| **Instructor-XL** | 93.30 | 83.85 | **86.76** | 77.50 | 75.47 | 69.46 | 74.39 | 46.22 | 56.94 | 60.35 | 67.43 | 56.78 | 57.58 |
| **SGPT** | 92.16 | 77.29 | 76.90 | 68.24 | 69.81 | 62.14 | 67.68 | 37.56 | 56.12 | 51.09 | 59.87 | 52.19 | 60.53 |
| **Falcon-7b** | 90.93 | 78.33 | 78.17 | 61.47 | 54.15 | 58.39 | 56.65 | 22.22 | 52.45 | 51.74 | 57.43 | 44.24 | 60.42 |
| | | | | | GEN_NLP_Reasoning | | | | | | | | |
| **GPT3.5** | 90.00 | 79.58 | 48.03 | 46.96 | 48.68 | 52.11 | 48.72 | 11.11 | **77.35** | 38.29 | 59.93 | 31.29 | 54.42 |
| **PaLM2** | 78.66 | 73.54 | 52.96 | 54.71 | 52.08 | 51.43 | 56.71 | 17.56 | 76.12 | 38.64 | 56.69 | 20.23 | **71.26** |
| **Dolly2-12b** | 62.27 | 59.05 | 18.86 | 20.74 | 13.27 | 34.91 | 46.59 | 24.00 | 62.24 | 25.58 | 41.91 | 24.53 | 21.79 |
| **Falcon-7b** | 54.43 | 7.71 | 1.27 | 0.00 | 1.70 | 8.57 | 14.45 | 11.11 | 72.24 | 2.31 | 25.30 | 26.66 | 13.47 |
| | | | | | RET_GEN_NLP_Reasoning | | | | | | | | |
| **GPT3.5** | 92.27 | **86.04** | 83.94 | **80.44** | 80.94 | **80.18** | 72.56 | 44.00 | 64.49 | **65.99** | **73.88** | **58.46** | 62.95 |
| **PaLM2** | 85.26 | 84.79 | 85.49 | 79.12 | 80.75 | 72.86 | 74.39 | 39.78 | 66.33 | 64.79 | 68.68 | 53.65 | 58.84 |
| **Dolly2-12b** | 91.55 | 85.00 | 85.21 | 76.76 | **84.53** | 71.07 | 71.95 | **48.67** | 56.73 | 63.19 | 69.47 | 55.74 | 36.95 |
| **Falcon-7b** | 90.10 | 78.33 | 78.17 | 61.47 | 54.15 | 60.18 | 58.41 | 24.44 | 62.04 | 52.03 | 57.43 | 49.90 | 50.84 |

Table 5: System Performance (Recall)%. The best result per medical variable is in bold.